\documentclass[lettersize,journal]{IEEEtran}

\usepackage{amsmath}
\usepackage{amssymb}
\usepackage{graphicx}
\usepackage{algorithm}
\usepackage{algorithmic}
\usepackage{url}
\usepackage{booktabs}    
\usepackage[table]{xcolor}

\title{\LARGE \bf
Efficient Visual Anomaly Detection at the Edge:\\ Enabling Real-Time Industrial Inspection on Resource-Constrained Devices

}

\author{Arianna Stropeni$^{1,*}$, Fabrizio Genilotti$^{1,*}$, Francesco Borsatti$^{1}$,\\ Manuel Barusco$^{1}$, Davide Dalle Pezze$^{1}$, Gian Antonio Susto$^{1}$%
\thanks{$^{1}$University of Padua, Padua, Italy}
\thanks{$^{*}$Authors contributed equally to this work}
}

\begin{document}

\maketitle
\thispagestyle{empty}
\pagestyle{empty}

%%%%%%%%%%%%%%%%%%%%%%%%%%%%%%%%%%%%%%%%%%%%%%%%%%%%%%%%%%%%%%%%%%%%%%%%%%%%%%%%
\begin{abstract}
Visual Anomaly Detection (VAD) is essential for industrial quality control, enabling automatic defect detection in manufacturing.
In real production lines, VAD systems must satisfy strict real-time and privacy requirements, necessitating a shift from cloud-based processing to local edge deployment.
However, processing data locally on edge devices introduces new challenges because edge hardware has limited memory and computational resources.
To overcome these limitations, we propose two efficient VAD methods designed for edge deployment: PatchCore-Lite and Padim-Lite, based on the popular PatchCore and PaDiM models.
PatchCore-Lite runs first a coarse search on a product-quantized memory bank, then an exact search on a decoded subset. 
Padim-Lite is sped up using diagonal covariance, turning Mahalanobis distance into efficient element-wise computation.
We evaluate our methods on the MVTec AD and VisA benchmarks and show their suitability for edge environments. 
PatchCore-Lite achieves a remarkable 79\% reduction in total memory footprint, while PaDiM-Lite achieves substantial efficiency gains with a 77\% reduction in total memory and a 31\% decrease in inference time.
These results show that VAD can be effectively deployed on edge devices, enabling real-time, private, and cost-efficient industrial inspection.
\end{abstract}

%%%%%%%%%%%%%%%%%%%%%%%%%%%%%%%%%%%%%%%%%%%%%%%%%%%%%%%%%%%%%%%%%%%%%%%%%%%%%%%%
\section{INTRODUCTION}

Visual Anomaly Detection (VAD) plays a critical role in modern manufacturing, enabling automated inspection of products to identify defects. Traditional VAD systems typically operate in centralized computing environments with high-performance hardware and processing images captured from production lines. 

The deployment of VAD systems at the edge means VAD models run on tiny devices located near the production line, with limited memory and computational resources.
This setting is highly relevant for industrial applications, as it offers several significant advantages.
First, processing images locally on edge devices eliminates the need to transmit potentially sensitive manufacturing data over the internet to cloud servers, thereby addressing privacy and intellectual property concerns. Second, edge processing enables lower latency inference, which is essential for fast-moving production lines where real-time decisions must be made about product quality. 

Despite these advantages, deploying VAD methods on edge devices introduces significant technical challenges. Edge devices typically have limited computation, memory and processing power.
Many state-of-the-art VAD methods have been developed with a primary focus on detection performance, often employing computationally expensive operations that are infeasible for edge deployment.

In this work, we proposed to modify two of the most well-known VAD models for the edge scenario: PatchCore \cite{roth2022towards} and PaDiM \cite{defard2021padim}.
While both methods achieve strong performance, they pose computational challenges for edge deployment: PaDiM can be computationally intensive due to the computation of covariance matrices, while PatchCore's memory bank can become prohibitively large on memory-constrained devices.

In this work, we address these challenges by proposing efficient variants of both methods specifically designed for edge deployment. 
Our contributions can be summarized as follows:

\begin{itemize}

\item We introduce a computationally efficient reformulation of PaDiM that replaces full-covariance estimation with a diagonal approximation, transforming the Mahalanobis distance into a lightweight, element-wise operation. This reduces the complexity from $O(d^3)$ to $O(d)$ for both the training and inference phases.

\item Our PatchCore variant employs a two-stage nearest-neighbour search: an initial coarse search is performed on a memory bank compressed via product quantisation, followed by an exact search on a small, selectively decoded subset. This achieves a significant storage reduction (12× reduction) and speed improvement.

\item The effectiveness of PatchCore-Lite and PaDiM-Lite is validated on MVTec and Visa datasets. Furthermore, resource-consumption metrics are provided to demonstrate the feasibility of deploying these models in edge scenarios.

\item For transparency and to support future research in visual anomaly detection, both methods are released publicly as part of the MoViAD \cite{barusco2025moviad} library.

\end{itemize}

The remainder of this paper is organized as follows. Section \ref{sec:related_work} reviews the related work in visual anomaly detection and edge computing. Section \ref{sec:methodology} presents our methodology contributions, including the mathematical formulations for PatchCore-Lite and PaDiM-Lite. Section \ref{sec:exp_setup} describes our experimental setup and the employed datasets. Section \ref{sec:results} presents experimental results, and Section \ref{sec:conclusion} draws some conclusions and possible future work directions.

\begin{figure}[t]
    \centering
    \includegraphics[width=\linewidth]{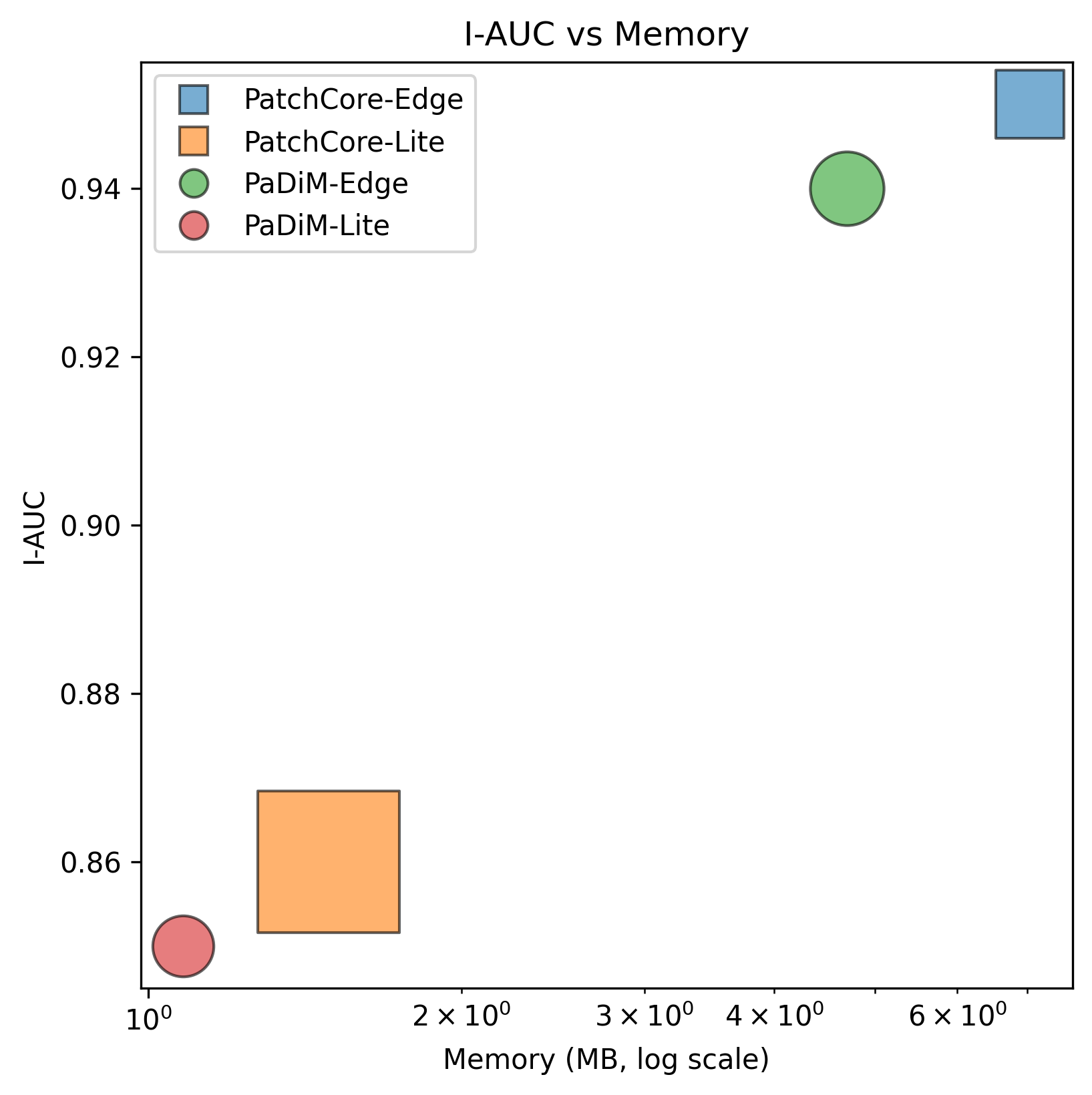}
    \caption{VAD models for the edge comparing Memory (log-scale), performance (I-AUC).}
    \label{fig:pipeline}
\end{figure}

\section{RELATED WORK}
\label{sec:related_work}

\subsection{Visual Anomaly Detection}

VAD plays a crucial role in many computer vision applications, including manufacturing, healthcare, autonomous driving, and security. VAD models offer two main advantages. First, they are typically trained in an unsupervised manner using only normal samples, bypassing the time-consuming and resource-intensive process of collecting and labeling large numbers of defective examples. Second, in addition to providing binary, sample-level predictions (normal vs. anomalous) to support decision-making, VAD methods also produce pixel-level anomaly maps, which enhance interpretability and increase end-user trust.

In particular, in the manufacturing domain, these properties are valuable because defective samples are scarce.
Therefore, supervised inspection systems require large amounts of labeled anomaly data, which is often impractical since defects are rare, costly to reproduce, and continuously evolving. 
By learning exclusively from normal production data, VAD systems can detect previously unseen defects. Furthermore, industrial inspection imposes strict requirements beyond simply flagging a product as defective; operators need localized anomaly maps to understand the root causes of failures. 

State-of-the-art VAD models generally fall into two main categories: reconstruction-based methods and feature embedding–based methods.\\
\textbf{Reconstruction-based methods} employ generative models to learn the distribution of normal data during training. At inference time, anomalies are identified through high reconstruction errors. Common approaches in this category include Autoencoders, Generative Adversarial Networks (GANs), and Diffusion Models.
However, these approaches are often computationally expensive, since they rely on complex generative models that must accurately reconstruct high-dimensional inputs. 
This can limit their applicability in real-time and resource-constrained scenarios such as edge deployment.\\
\textbf{Feature embedding–based methods}, in contrast, exploit representations extracted from pretrained neural networks, avoiding explicit image reconstruction and generally achieving higher computational efficiency than reconstruction-based approaches. This family of methods can be further organized into three main subcategories.
\textbf{Teacher–Student methods} utilize knowledge distillation between a teacher and a student network, where discrepancies between their feature maps signal anomalous regions.
\textbf{Memory Bank methods} store feature representations of normal samples in a memory bank; approaches such as PaDiM \cite{defard2021padim}, PatchCore \cite{roth2022towards}, and CFA \cite{9839549} belong to this category.
\textbf{Normalizing Flow methods} use flow-based models to map complex data distributions to a normal distribution, enabling anomaly detection through likelihood estimation.

\subsection{Edge-Based Visual Anomaly Detection}
In practical scenarios, such as manufacturing, VAD models are often required to operate close to the data source, which necessitates deployment on edge devices. 
Despite the importance of this setting, only a limited number of works have addressed model optimization for edge deployment.
In particular, \cite{barusco2025paste} presents the first benchmark for deploying VAD models on the edge. 
They replace standard heavy backbones, such as WideResNet50, with lightweight architectures like MobileNetV2 in major feature-based VAD methods, and evaluate them on the most widely used benchmarks in the field, namely MVTec AD and VisA.
This straightforward approach achieves substantial reductions in memory footprint and computational requirements while maintaining competitive detection performance. 
Building on these findings, the authors introduce PaSTe, an edge-optimized VAD model based on the STFPM architecture \cite{wang2021studentteacherfeaturepyramidmatching}, which establishes a strong baseline for future research in resource-constrained anomaly detection.

While \cite{barusco2025paste} establishes PaSTe as the first architectural adaptation for edge-based VAD, their work focuses exclusively on a method, STFPM, that belongs to the teacher-student category. 
In this context, we propose two novel efficient VAD methods designed for edge deployment, both belonging to the memory bank category, whose adaptation for edge scenarios remains unexplored. 
Specifically, starting with a tiny backbone as suggested in \cite{barusco2025paste}, we further optimise PatchCore \cite{roth2022towards}, and PaDiM \cite{defard2021padim} by introducing PatchCore-Lite and PaDiM-Lite.

Respectively, we propose specific design modifications enabling efficient execution on resource-constrained devices.

\section{METHODOLOGY}
\label{sec:methodology}

This section presents our two main contributions: a simplified version of PaDiM using variance vectors (PaDiM-Lite) and an improved PatchCore with product quantization for memory bank compression (PatchCore-Lite).

\subsection{Backbone}
Both PaDiM and PatchCore rely on a frozen backbone to extract rich and discriminative feature representations from the input samples. 
The backbone is typically a large-scale architecture, such as WideResNet50, pretrained on ImageNet.
Specifically, features are collected from multiple intermediate layers of the network to capture information at different spatial resolutions and semantic levels, enabling the detection of both low-level texture anomalies and high-level structural defects. 
Specifically, given the feature extractor $f(\boldsymbol{x})$, where $\boldsymbol{x} \in \mathbb{R}^{H \times W \times C}$, the features extracted from different layers of the network are defined as: $\boldsymbol{f}_i \in \mathbb{R} ^ {H_i \times W_i \times d_i}$, where $i \in \{1, N\}$ is the layer index.
These multi-scale embeddings are reshaped to match their spatial resolutions and concatenated along the channel dimension in order to form a unique feature map $\boldsymbol{F} \in \mathbb{R}^{H^* \times W^* \times d}$. The set of $\boldsymbol{x}_{i,j} \in F$ is the set of patch-level descriptors, which serve as the basis for subsequent anomaly scoring.

Since the backbone dominates both memory consumption and inference latency, its architecture plays a crucial role when deploying VAD systems on resource-constrained edge devices. Lightweight backbones, such as MobileNet, significantly reduce computational complexity and memory footprint while maintaining sufficient representational power. 
In the results section, we refer to the methods with the light backbone as PatchCore-Edge and PaDiM-Edge, while our methodological adaptations will be called PatchCore-Lite and PaDiM-Lite, respectively.

\subsection{PaDiM-Lite}

\subsubsection{PaDiM}
PaDiM models the distribution of normal patch descriptors using multivariate Gaussian distributions \cite{defard2021padim}. For each spatial position $(i,j)$ in the feature map $F$, obtained with a pretrained feature extractor (in our case MobileNetV2), PaDiM estimates a Gaussian distribution $\mathcal{N}(\boldsymbol{\mu}_{ij}, \boldsymbol{\Sigma}_{ij})$ from the training features, where $\boldsymbol{\mu}_{ij} \in \mathbb{R}^d$ is the mean vector and $\boldsymbol{\Sigma}_{ij} \in \mathbb{R}^{d \times d}$ is the full covariance matrix.

At test time, the anomaly score for a patch at position $(i,j)$ is computed using the Mahalanobis distance:

\begin{equation}
M(\mathbf{x}_{ij}) = (\mathbf{x}_{ij} - \boldsymbol{\mu}_{ij})^T \boldsymbol{\Sigma}_{ij}^{-1} (\mathbf{x}_{ij} - \boldsymbol{\mu}_{ij})
\end{equation}

where $\mathbf{x}_{ij} \in \mathbb{R}^d$ is the feature vector at position $(i,j)$ for the test image.

\textbf{Computational Complexity of Original PaDiM:}

\textit{Training Phase:}
\begin{itemize}
\item Computing the full covariance matrix: $O(d^2 \cdot N)$ 
\item Inverting the covariance matrix: $O(d^3)$ (typically using Cholesky decomposition)
\item Total training complexity per position: $O(d^2 \cdot N + d^3)$
\end{itemize}
where $d$ is the embedding dimension and $N$ the number of samples.

\textit{Inference Phase:}
\begin{itemize}
\item Matrix-vector multiplication $\boldsymbol{\Sigma}_{ij}^{-1} (\mathbf{x}_{ij} - \boldsymbol{\mu}_{ij})$: $O(d^2)$
\item Final dot product: $O(d)$
\item Total inference complexity per patch: $O(d^2)$
\end{itemize}

\subsubsection{PaDiM-Lite}
We propose a computationally efficient reformulation of PaDiM that replaces the full covariance estimation by approximating it with a diagonal covariance matrix, effectively using only the variance vector.
This is motivated by the observation that while feature correlations can improve modeling accuracy, they come at a high computational cost that may not be justified for edge deployment, where computational resources are limited.

Specifically, we assume independence between feature dimensions and model each position $(i,j)$ using:

\begin{equation}
\boldsymbol{\sigma}_{ij}^2 = \text{diag}(\boldsymbol{\Sigma}_{ij} )  = (\sigma_{ij,1}^2, \sigma_{ij,2}^2, \ldots, \sigma_{ij,d}^2) 
\end{equation}

where $\boldsymbol{\sigma}_{ij}^2 \in \mathbb{R}^d$ is the variance vector with elements:

\begin{equation}
\sigma_{ij,k}^2 = \frac{1}{N-1} \sum_{n=1}^{N} (x_{ij,k}^{(n)} - \mu_{ij,k})^2
\end{equation}

For numerical stability, we add a small regularization term $\epsilon$ (typically $\epsilon = 0.01$):

\begin{equation}
\sigma_{ij,k}^2 = \sigma_{ij,k}^2 + \epsilon
\end{equation}

The Mahalanobis distance simplifies to:

\begin{equation}
M(\mathbf{x}_{ij}) = \sum_{k=1}^{d} \frac{(x_{ij,k} - \mu_{ij,k})^2}{\sigma_{ij,k}^2}
\end{equation}

\textbf{Complexity Analysis:}

\textit{Training Phase:}
\begin{itemize}
\item Computing variance vector: $O(d \cdot N)$
\item No matrix inversion needed
\item Total training complexity per position: $O(d \cdot N)$
\item \textbf{Reduction from $O(d^2 \cdot N + d^3)$ to $O(d \cdot N)$}
\end{itemize}

\textit{Inference Phase:}
\begin{itemize}
\item Element-wise squared differences: $O(d)$
\item Element-wise divisions: $O(d)$
\item Summation: $O(d)$
\item Total inference complexity per position: $O(d)$
\item \textbf{Reduction from $O(d^2)$ to $O(d)$}
\end{itemize}

where $d$ represents the embedding dimension of the vector and $N$ the number of samples.

\textbf{Memory Footprint:} This simplified approach also reduces memory requirements. Instead of storing a $d \times d$ matrix requiring $4 \times d^2$ bytes (for 32-bit floats), we store only a $d$-dimensional variance vector requiring $4 \times d$ bytes per position, plus the mean vector.

\subsection{PatchCore-Lite}

\subsubsection{PatchCore}
PatchCore~\cite{roth2022towards} maintains a memory bank $\mathcal{M} = \{\mathbf{m}_1, \mathbf{m}_2, \ldots, \mathbf{m}_K\}$ of patch feature vectors extracted from normal training images, where each $\mathbf{m}_i \in \mathbb{R}^d$. At test time, the anomaly score for a test patch feature $\mathbf{x}$ is computed as the distance to its nearest neighbor in the memory bank:

\begin{equation}
s(\mathbf{x}) = \min_{\mathbf{m} \in \mathcal{M}} \|\mathbf{x} - \mathbf{m}\|_2,
\end{equation}

while the anomaly score of the entire image is calculated as the maximum distance score between the patch feature vectors of the input image and each feature vector $\ \boldsymbol{m}_{i} \in \mathcal{M}$ subsequently regularized.

While effective for anomaly detection, this approach presents significant challenges for edge deployment. For a memory bank of size $K$ (typically 10,000 samples after coreset selection) with $d$-dimensional features stored as 32-bit floats, the memory requirement is $4\times K\times d$ bytes. Moreover, the inference phase requires $K$ distance computations per test patch to find the nearest neighbor, which can be computationally expensive.

\subsubsection{PatchCoreLite: Memory Bank with Product Quantization}
We propose to compress the PatchCore memory bank using product quantization, which is applied after standard PatchCore training. The process consists of the following steps:

\textbf{Step 1: Standard PatchCore Training}. We first train PatchCore normally by extracting features from the training set and performing coreset selection to build the memory bank $\mathcal{M}$ of size $K$.

\textbf{Step 2: Product Quantization}. We apply product quantization to compress each feature vector in the memory bank. The $d$-dimensional space is decomposed into $m$ subspaces, and each feature vector $\mathbf{m} \in \mathbb{R}^d$ is split into $m$ sub-vectors:

\begin{equation}
\mathbf{m} = [\mathbf{m}^{(1)}, \mathbf{m}^{(2)}, \ldots, \mathbf{m}^{(m)}]
\end{equation}

where each $\mathbf{m}^{(j)} \in \mathbb{R}^{d/m}$. For each subspace $j \in [1,m]$, we learn a codebook $\mathcal{C}_j = \{\mathbf{c}_j^1, \mathbf{c}_j^2, \ldots, \mathbf{c}_j^{V}\}$ of $V$ centroids using k-means clustering on the corresponding sub-vectors from all memory bank features. Typically, $V = 2^b$ for $b$ bits per subspace (e.g., $b = 8$ gives 256 centroids).

Each sub-vector $\mathbf{m}^{(j)}$ is then quantized by finding its nearest centroid:

\begin{equation}
q_j(\mathbf{m}) = \arg\min_{i \in [1,V]} \|\mathbf{m}^{(j)} - \mathbf{c}_j^i\|_2
\end{equation}

The quantized representation of $\mathbf{m}$ consists of $m$ indices $[q_1(\mathbf{m}), q_2(\mathbf{m}), \ldots, q_m(\mathbf{m})]$, requiring only $m \times b$ bits instead of $32 \times d$ bits for the original vector.

\textbf{Step 3: Storage}. The compressed memory bank is stored on the edge device as:
\begin{itemize}
\item Quantization indices for all $K$ memory vectors: $(m \times b \times K) / 8$ bytes.
\item Learned codebooks for all $m$ subspaces: $4 \times m \times V \times (d/m) = 4 \times V \times d$ bytes.
\end{itemize}

For typical values ($K=10000$, $d=256$, $m=8$, $b=8$), the quantized indices require 78.12 KB and the codebooks further 256 KB, resulting in a total storage of approximately 334 KB compared to the original 9.77 MB.

\subsubsection{Inference Phase: Two-Stage Nearest Neighbor Search}

At inference time, we employ an efficient two-stage search strategy that balances speed and accuracy:

\textbf{Step 1: Coarse Search in Compressed Space}. For a test patch feature $\mathbf{x} \in \mathbb{R}^d$, we first quantize it using the learned codebooks:

\begin{equation}
\mathbf{ \tilde{x}} = [q_1(\mathbf{x}^{(1)}), q_2(\mathbf{x}^{(2)}), \ldots, q_{\mathbf{m}}(\mathbf{x}^{(\mathbf{m})})]
\end{equation}

We compute the asymmetric distance to each quantized memory vector using:

\begin{equation}
\hat{d}(\mathbf{x}, \mathbf{m}_i) = \sqrt{\sum_{j=1}^{m} \|\mathbf{\tilde{x}_{j}} - \mathbf{c}_j^{q_j(\mathbf{m}_i)}\|_2^2}
\end{equation}

We then identify the $k$ nearest neighbors in the compressed space, where $k$ is an hyperparameter (typically $k = 500$ to $1000$ out of $K = 10000$) where $k \ll K$ .

\textbf{Step 2: Fine Search on Decoded Subset}. We decode only the $k$ candidate nearest neighbors identified in Step 1. For each quantized memory vector $\mathbf{m}_i$ in the candidate set, we reconstruct an approximation:

\begin{equation}
\tilde{\mathbf{m}}_i = [\mathbf{c}_1^{q_1(\mathbf{m}_i)}, \mathbf{c}_2^{q_2(\mathbf{m}_i)}, \ldots, \mathbf{c}_m^{q_m(\mathbf{m}_i)}]
\end{equation}

We then compute the exact Euclidean distance between $\mathbf{x}$ and each decoded vector $\tilde{\mathbf{m}}_i$ in the candidate set, and select the minimum as the final anomaly score:

\begin{equation}
s(\mathbf{x}) = \min_{\tilde{\mathbf{m}}_i \in \text{candidates}} \|\mathbf{x} - \tilde{\mathbf{m}}_i\|_2
\end{equation}

\textbf{Complexity Analysis}. The two-stage approach provides significant computational savings:
\begin{itemize}
\item \textbf{Stage 1}: $O(K \cdot m)$ operations for approximate distance computation using lookup tables, where $m$ is typically small (e.g., 8)
\item \textbf{Stage 2}: $O(k \cdot d)$ operations for exact distance computation on the small candidate set, where $k \ll K$
\item \textbf{Total}: $O(K \cdot m + k \cdot d)$ compared to $O(K \cdot d)$ for exhaustive search
\end{itemize}

For typical values ($K=10000$, $k=1000$, $d=256$, $m=8$), this represents a significant reduction in computations. Furthermore, to reduce the number of comparisons needed to calcolate the image level anomaly score in the original PatchCore implementation, we consider as image level anomaly score the maximum patch anomaly score in the image (as done also in PaDiM).

\begin{figure}[t]
    \centering
    \includegraphics[width=0.9\linewidth]{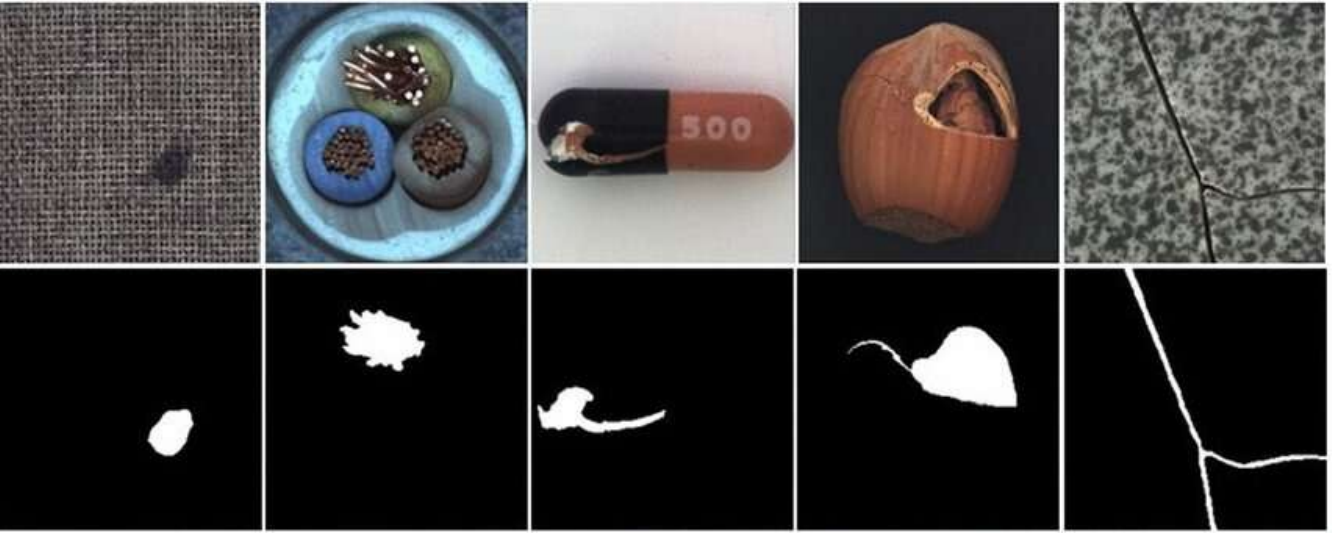}
    \caption{Examples of anomalous images present in the test dataset of the MVTeC \cite{bergmann2019mvtec} with their corresponding anomaly mask.}
    \label{fig:examples_dataset}
\end{figure}

\section{EXPERIMENTAL SETUP}
\label{sec:exp_setup}

\subsection{Datasets}
For evaluation, we make use of two well-established benchmarks in the Visual Anomaly Detection field: MVTec and Visa datasets (see Figure \ref{fig:examples_dataset} for representative examples). 
The MVTec AD dataset \cite{bergmann2019mvtec} consists of high-resolution images across 15 categories, including 10 object types (bottle, hazelnut, metal nut, etc.) and 5 texture types (grid, tiles, etc.). Each category provides only defect-free training images, while the test set contains both normal and anomalous samples with pixel-level ground truth annotations available for all anomalous test images.
Similarly, the VisA dataset \cite{zou2022spotthedifferenceselfsupervisedpretraininganomaly} comprises 10,821 high-resolution color images (9,621 normal and 1,200
anomalous samples) covering 12 objects in 3 domains, making it another recognised benchmark for industrial anomaly detection.

\subsection{Implementation Details}

For both proposed methods, we employ a pre-trained MobileNetV2 backbone as the feature extractor, which provides an optimal balance between representation power and computational efficiency for edge deployment, as demonstrated in \cite{barusco2025paste}. Features are extracted from three intermediate layers to capture multi-scale semantic information at different spatial resolutions. The complete implementation is available in the open-source MoViAD library \cite{barusco2025moviad}.

Both PaDiM-Edge and PaDiM-Lite extract features from layers 7, 10, and 13 of MobileNetV2, resulting in a 224-dimensional concatenated feature vector. 
For PatchCore-Edge and PatchCore-Lite, we follow the implementation of \cite{barusco2025paste} and extract features from layers 4, 7, and 10 of MobileNetV2. The product quantization in PatchCore-Lite uses $m=8$ subspaces with $b=8$ bits per subspace (256 centroids per subspace).
For STFPM-Edge and Paste methods, we use the layers 3, 8, and 14 for feature extraction as proposed in the original work \cite{barusco2025paste}. For both methods, the teacher is the backbone, and the student is the additional memory.

% % versione del 7 febbraio 
% \begin{table}[t]
% \centering
% \caption{Performance comparison on MVTec and VisA datasets.}
% \begin{tabular}{|c|c|c|c|c|}
% \hline
% Method & \multicolumn{2}{c|}{I-ROC} & \multicolumn{2}{c|}{P-ROC} \\ \hline
%        & MVTec & VisA & MVTec & VisA \\ \hline
% PatchCore  \cite{roth2022towards}    & 0.98 & 0.92 & 0.97 & 0.98 \\ \hline
% PatchCore-Edge \cite{barusco2025paste}  & 0.95 & 0.86 & 0.96 & 0.91 \\ \hline
% PatchCore-Lite  & 0.86 & 0.68 & 0.94 & 0.90 \\ \hline \hline
% PaDiM   \cite{defard2021padim}        & 0.96    & 0.88 &  0.98   & 0.94 \\ \hline
% PaDiM-Edge \cite{barusco2025paste}      & 0.94 & 0.84 & 0.97 & 0.97 \\ \hline
% PaDiM-Lite      & 0.85 & 0.75 & 0.94 & 0.93 \\ \hline \hline
% STFPM    \cite{wang2021studentteacherfeaturepyramidmatching}       & 0.95 & 0.85 & 0.97 & 0.95 \\ \hline
% STFPM-Edge  \cite{barusco2025paste}   & 0.90 & 0.84 & 0.96 & 0.97 \\ \hline
% Paste   \cite{barusco2025paste}  & 0.92 & 0.84 & 0.95 & 0.97 \\ \hline
% \end{tabular}
% \end{table}

\begin{table}[ht!]
    \caption{Performance comparison on MVTec and VisA datasets.}
    \label{tab:performance_comparison}
    \centering
    % \rowcolors{2}{lightgray}{}
    \begin{tabular}{lcccc}
        \midrule
        \textbf{Method} & \multicolumn{2}{c}{\textbf{I-ROC}} & \multicolumn{2}{c}{\textbf{P-ROC}} \\
        \cmidrule(lr){2-3} \cmidrule(lr){4-5}
         & \textbf{MVTec} & \textbf{VisA} & \textbf{MVTec} & \textbf{VisA} \\
        \midrule
        PatchCore \cite{roth2022towards} & 0.98 & 0.92 & 0.97 & 0.98 \\
        PatchCore-Edge \cite{barusco2025paste} & 0.95 & 0.86 & 0.96 & 0.91 \\
        PatchCore-Lite & 0.86 & 0.68 & 0.94 & 0.90 \\
        \midrule
        PaDiM \cite{defard2021padim} & 0.96 & 0.88 & 0.98 & 0.94 \\
        PaDiM-Edge \cite{barusco2025paste} & 0.94 & 0.84 & 0.97 & 0.97 \\
        PaDiM-Lite & 0.85 & 0.75 & 0.94 & 0.93 \\
        \midrule
        STFPM \cite{wang2021studentteacherfeaturepyramidmatching} & 0.95 & 0.85 & 0.97 & 0.95 \\
        STFPM-Edge \cite{barusco2025paste} & 0.90 & 0.84 & 0.96 & 0.97 \\
        Paste \cite{barusco2025paste} & 0.92 & 0.84 & 0.95 & 0.97 \\
        \midrule
    \end{tabular}
    \vspace{-1em}
\end{table}

\begin{table*}[t]
    \caption{Efficiency comparison across models.}
    \centering
    % \rowcolors{2}{lightgray}{}
    \begin{tabular}{lccccc}
    \midrule
    \textbf{Method} & \textbf{Backbone [MB]} & \textbf{Additional Memory [MB]} & \textbf{Total Memory [MB]} & \textbf{Inf. Time [ms]} & \textbf{Peak Memory Inference [MB]} \\
    \midrule
    PatchCore \cite{roth2022towards} & 95 & 68 & 163 & 345 & 286 \\
    PatchCore-Edge \cite{barusco2025paste} & 0.95 & 6.10 & 7.05 & 30 & 0.02 \\
    PatchCore-Lite (ours) & 0.95 & 0.54 & 1.49 & 130 & 0.03 \\
    \midrule
    PaDiM \cite{defard2021padim} & 95 & 3800 & 3895 & 22553 & 1.26 \\
    PaDiM-Edge \cite{barusco2025paste} & 0.95 & 3.75 & 4.70 & 35 & 0.91 \\
    PaDiM-Lite (ours) & 0.95 & 0.13 & 1.08 & 24 & 0.68 \\
    \midrule
    STFPM \cite{wang2021studentteacherfeaturepyramidmatching} & 95 & 95 & 190 & 123 & 332 \\
    STFPM-Edge \cite{barusco2025paste} & 2.66 & 2.66 & 5.32 & 13 & 18.34 \\
    Paste \cite{barusco2025paste} & 2.66 & 2.45 & 5.11 & 9 & 44.62 \\
    \midrule
    \end{tabular}
    \label{tab:efficiency_comparison}
\end{table*}

\subsection{Evaluation Metrics}

We evaluate both methods using standard metrics for visual anomaly detection:

\begin{itemize}
\item \textbf{Image-level AUROC (I-ROC)}: Area under the Receiver Operating Characteristic Curve for classifying images as normal or anomalous.
\item \textbf{Pixel-level AUROC (P-ROC)}: Area under the Receiver Operating Characteristic Curve for classifying each pixel as normal or anomalous.
\item \textbf{Memory Footprint}: Total memory required to store model parameters and auxiliary data structures (memory bank, statistics, etc.).
\item \textbf{Inference Time}: Average time required to process a single test image on a target edge device.
\item \textbf{Inference Peak memory}: Average peak memory needed to process a single test image on a target edge device.
\end{itemize}

All experiments to measure the Inference Time and Peak Memory Inference metrics are conducted on a Intel(R) Core(TM) i5-9600K CPU @ 3.70GHz to simulate edge deployment conditions.

\section{EXPERIMENTAL RESULTS}
\label{sec:results}

Section \ref{subsec:padim_comparison} presents a detailed comparison of PaDiM-Lite and PaDiM-Edge in terms of memory footprint and computational efficiency. Section \ref{subsec:patchcore_comparison} provides an analogous analysis for PatchCore-Lite and PatchCore-Edge.

\subsection{Comparison of Padim-Lite vs. Padim-Edge}
\label{subsec:padim_comparison}

\textbf{Memory and Inference Comparison: }
PaDiM-Lite by replacing the full covariance matrix estimation with a diagonal approximation, achieves a significant reduction in the additional memory, moving from 3.75 MB to 0.13 MB (97\%), while total memory consumption (including the MobileNetV2 backbone) decreases by 77\%. 
Inference latency improves by 31\%, decreasing from 35 ms to 24 ms per image. 
This speedup reflects the theoretical complexity reduction from $O(d^2)$ to $O(d)$ for the Mahalanobis distance computation, though due to the time spent by the feature extraction step, the total inference time is reduced just by 31\%.
\\
\textbf{Performance: }These reductions are achieved while maintaining strong pixel-level localization performance, with P-ROC values of 0.94 on MVTec AD compared to 0.97 of Padim-Edge, though there is a slight decrease for image-level performance, from 0.94 of Padim-Edge to 0.85 of Padim-Lite.
This demonstrates that the diagonal approximation preserves the essential discriminative information needed for accurate anomaly localization, discarding the cross-dimensional correlations that, while theoretically informative, prove less critical for spatial localization and come at a prohibitive computational cost for edge scenarios.

\subsection{Comparison of PatchCore-Lite vs. PatchCore-Edge}
\label{subsec:patchcore_comparison}

\textbf{Memory and Inference Comparison: } The quantization of PatchCore’s memory bank enables a 12× reduction in memory compared to PatchCore-Edge, which itself already reduced the memory footprint of the original PatchCore by 12× in the memory bank and 23× overall. On the other hand, inference time is approximately four times higher than PatchCore-Edge, due to the dequantization performed on the $k$ nearest neighbors during the second step of inference. While this is computationally lighter than dequantizing the entire memory bank, it still leads to a noticeable increase in inference time. Nevertheless, the peak memory consumption at inference remains low, and combined with the minimal storage requirements, this makes the approach particularly well-suited for memory-constrained devices.
PatchCore-Lite has the most flexible architecture in terms of memory usage, since both the number of samples in the memory bank and the quantization hyperparameters can be tuned to the exact hardware specifications of the deployed edge device. This allows to get the most out of the available hardware resources and tune the performance/efficiency trade-off.

\textbf{Performance: }
Employing PatchCore-Lite results in a reduction in image-level performance from 0.95 to 0.86, which is expected, as quantization reduces the representational capacity of the memory bank. Nevertheless, the degradation is limited, and even negligible at pixel-level. A similar trend is observed on the VisA dataset, where the drop in image-level performance is more pronounced. This effect is likely due to the nature of the anomalies present in the VisA dataset, for which image-level scores may be more sensitive to quantization errors. Finally, it is worth noting that performance is highly sensitive to the number of nearest neighbors ($k$) that are chosen: typically, the higher $k$ is, the higher will be the performance, at the expense of an increased inference time related to the need of dequantizing a higher number of vectors.

\section{CONCLUSIONS}
\label{sec:conclusion}

In this work, we addressed the critical challenge of deploying Visual Anomaly Detection systems on resource-constrained edge devices for real-time industrial inspection. We proposed two efficient VAD methods, PatchCore-Lite and PaDiM-Lite, specifically designed to overcome the memory and computational limitations inherent in edge environments while maintaining practical detection performance.

PaDiM-Lite achieves substantial efficiency gains by replacing full covariance matrix estimation with a diagonal approximation, resulting in a 77\% total memory reduction compared to PaDiM-Edge and a 31\% reduction in inference time, while maintaining competitive pixel-level performance (P-ROC of 0.94 on MVTec AD compared to 0.97 of Padim-Edge).
PatchCore-Lite employs a two-stage nearest-neighbor search strategy with product quantization, achieving a remarkable 79\% total memory reduction compared to PatchCore-Edge and a ~91\% reduction when considering only the memory bank.
This compression comes at the cost of a 4× increase in inference time (130 ms vs. 30 ms), primarily due to the selective dequantization of candidate neighbors during the fine search stage.

Our experimental evaluation on the MVTec AD and VisA benchmarks demonstrates that both methods achieve a favorable trade-off between detection performance and resource efficiency. These results confirm that effective VAD can be deployed on edge devices, enabling real-time, privacy-preserving, and cost-efficient industrial quality control without reliance on cloud infrastructure.

\bibliographystyle{IEEEtran}  
\bibliography{main}

% Generated by IEEEtran.bst, version: 1.14 (2015/08/26)
\begin{thebibliography}{1}
\providecommand{\url}[1]{#1}
\csname url@samestyle\endcsname
\providecommand{\newblock}{\relax}
\providecommand{\bibinfo}[2]{#2}
\providecommand{\BIBentrySTDinterwordspacing}{\spaceskip=0pt\relax}
\providecommand{\BIBentryALTinterwordstretchfactor}{4}
\providecommand{\BIBentryALTinterwordspacing}{\spaceskip=\fontdimen2\font plus
\BIBentryALTinterwordstretchfactor\fontdimen3\font minus \fontdimen4\font\relax}
\providecommand{\BIBforeignlanguage}[2]{{%
\expandafter\ifx\csname l@#1\endcsname\relax
\typeout{** WARNING: IEEEtran.bst: No hyphenation pattern has been}%
\typeout{** loaded for the language `#1'. Using the pattern for}%
\typeout{** the default language instead.}%
\else
\language=\csname l@#1\endcsname
\fi
#2}}
\providecommand{\BIBdecl}{\relax}
\BIBdecl

\bibitem{roth2022towards}
K.~Roth, L.~Pemula, J.~Zepeda, B.~Sch{\"o}lkopf, T.~Brox, and P.~Gehler, ``Towards total recall in industrial anomaly detection,'' in \emph{Proceedings of the IEEE/CVF conference on computer vision and pattern recognition}, 2022, pp. 14\,318--14\,328.

\bibitem{defard2021padim}
T.~Defard, A.~Setkov, A.~Loesch, and R.~Audigier, ``Padim: a patch distribution modeling framework for anomaly detection and localization,'' in \emph{International conference on pattern recognition}.\hskip 1em plus 0.5em minus 0.4em\relax Springer, 2021, pp. 475--489.

\bibitem{barusco2025moviad}
M.~Barusco, F.~Borsatti, A.~Stropeni, D.~D. Pezze, and G.~A. Susto, ``Moviad: A modular library for visual anomaly detection,'' \emph{arXiv preprint arXiv:2507.12049}, 2025.

\bibitem{9839549}
S.~Lee, S.~Lee, and B.~C. Song, ``Cfa: Coupled-hypersphere-based feature adaptation for target-oriented anomaly localization,'' \emph{IEEE Access}, vol.~10, pp. 78\,446--78\,454, 2022.

\bibitem{barusco2025paste}
M.~Barusco, F.~Borsatti, D.~D. Pezze, F.~Paissan, E.~Farella, and G.~A. Susto, ``Paste: Improving the efficiency of visual anomaly detection at the edge,'' in \emph{Proceedings of the Computer Vision and Pattern Recognition Conference}, 2025, pp. 4026--4035.

\bibitem{wang2021studentteacherfeaturepyramidmatching}
\BIBentryALTinterwordspacing
G.~Wang, S.~Han, E.~Ding, and D.~Huang, ``Student-teacher feature pyramid matching for anomaly detection,'' 2021. [Online]. Available: \url{https://arxiv.org/abs/2103.04257}
\BIBentrySTDinterwordspacing

\bibitem{bergmann2019mvtec}
P.~Bergmann, M.~Fauser, D.~Sattlegger, and C.~Steger, ``Mvtec ad--a comprehensive real-world dataset for unsupervised anomaly detection,'' in \emph{Proceedings of the IEEE/CVF conference on computer vision and pattern recognition}, 2019, pp. 9592--9600.

\bibitem{zou2022spotthedifferenceselfsupervisedpretraininganomaly}
\BIBentryALTinterwordspacing
Y.~Zou, J.~Jeong, L.~Pemula, D.~Zhang, and O.~Dabeer, ``Spot-the-difference self-supervised pre-training for anomaly detection and segmentation,'' 2022. [Online]. Available: \url{https://arxiv.org/abs/2207.14315}
\BIBentrySTDinterwordspacing

\end{thebibliography}

\end{document}